\documentclass[10pt,twocolumn,letterpaper]{article}

\usepackage{cvpr}
\usepackage{times}
\usepackage{epsfig}
\usepackage{graphicx}
\usepackage{amsmath}
\usepackage{amssymb}

\usepackage{multirow}
\usepackage{booktabs}
\usepackage{subfig}
\usepackage[ruled,noresetcount]{algorithm2e}
\usepackage[pagebackref=true,breaklinks=true,letterpaper=true,colorlinks,bookmarks=false]{hyperref}

\cvprfinalcopy 

\def\cvprPaperID{805} 

\ifcvprfinal\fi
\begin{document}

\title{Self-Correction for Human Parsing}

\author{
Peike Li\textsuperscript{\rm 1}, Yunqiu Xu\textsuperscript{\rm 2}, Yunchao Wei\textsuperscript{\rm 1}, Yi Yang\textsuperscript{\rm 1,2}\\ 
\textsuperscript{\rm 1}ReLER Lab, Centre for Artificial Intelligence, University of Technology Sydney\\
\textsuperscript{\rm 2}Baidu Research\\
\tt\small{peike.li@yahoo.com, imyunqiuxu@gmail.com, \{yunchao.wei, yi.yang\}@uts.edu.au} 
}

\maketitle

\begin{abstract}
Labeling pixel-level masks for fine-grained semantic segmentation tasks, \textit{e.g.} human parsing, remains a challenging task. The ambiguous boundary between different semantic parts and those categories with similar appearance usually are confusing, leading to unexpected noises in ground truth masks. To tackle the problem of learning with label noises, this work introduces a purification strategy, called Self-Correction for Human Parsing (SCHP), to progressively promote the reliability of the supervised labels as well as the learned models. In particular, starting from a model trained with inaccurate annotations as initialization, we design a cyclically learning scheduler to infer more reliable pseudo-masks by iteratively aggregating the current learned model with the former optimal one in an online manner. Besides, those correspondingly corrected labels can in turn to further boost the model performance. In this way, the models and the labels will reciprocally become more robust and accurate during the self-correction learning cycles. Benefiting from the superiority of SCHP, we achieve the best performance on two popular single-person human parsing benchmarks, including LIP and Pascal-Person-Part datasets. Our overall system ranks 1st in CVPR2019 LIP Challenge. Code is available at \href{https://github.com/PeikeLi/Self-Correction-Human-Parsing}{this url}.
\end{abstract}

\begin{figure}[t]
\centering
\subfloat[]{
\includegraphics[width=0.9\linewidth]{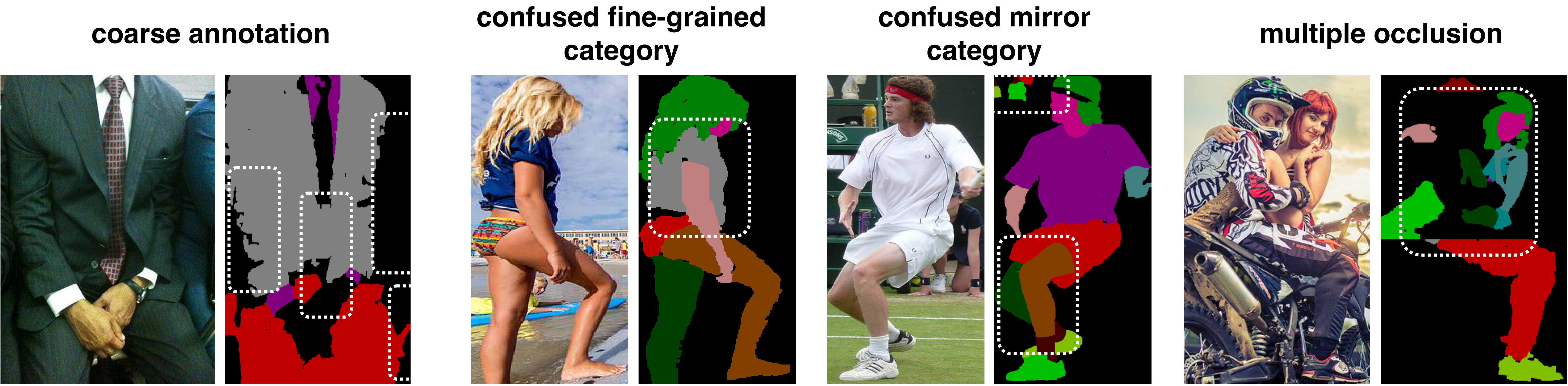}\label{fig:label-noise} 
}
\\
\subfloat[]{
\includegraphics[width=0.9\linewidth]{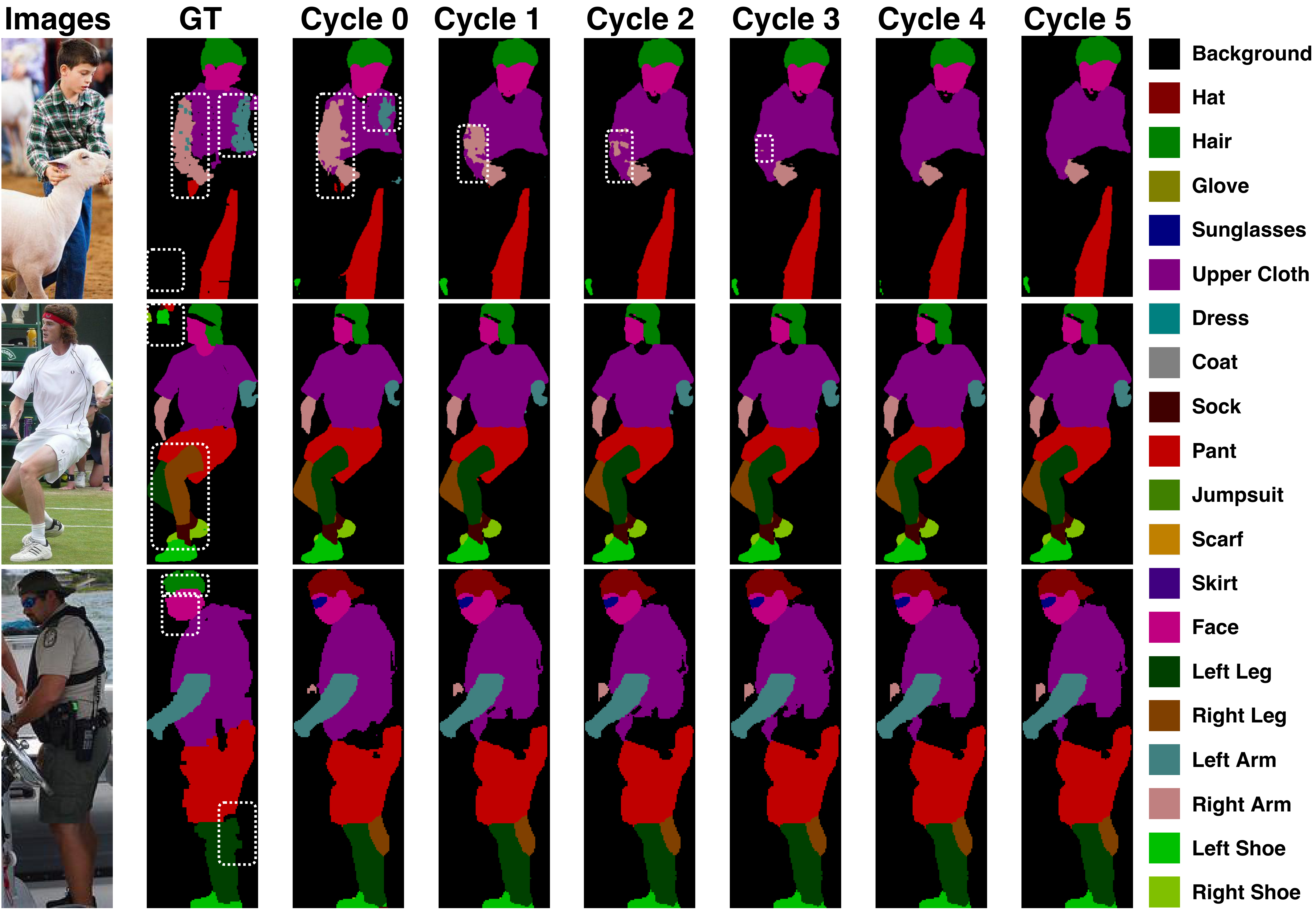}\label{fig:lip-self-correction} 
}  
\caption{(a) Different types of label noises exist in the ground truth. (b) Our self-correction mechanism progressively promotes the reliability of the supervised labels. Label noises are emphasized by white dotted boxes. Better zoom in to see the details.}
\end{figure} 
\section{Introduction}

Human parsing, as a fine-grained semantic segmentation task, aims to assign each image pixel from the human body to a semantic category, \textit{e.g.} arm, leg, dress, skirt. Understanding the detailed semantic parts of human is crucial in several potential application scenarios, including image editing, human analysis, virtual try-on and virtual reality. Recent advances on fully convolutional neural networks~\cite{long2015fully,chen2017deeplab} achieves various of well-performing methods for the human parsing task~\cite{liang2018look,ruan2019devil}.

To learn reliable models for human parsing, a large amount of pixel-level masks are required for supervision. However, labeling pixel-level annotations for human parsing is much harder than those traditional pixel-level understanding tasks. In particular, for those traditional semantic segmentation tasks~\cite{long2015fully,chen2017deeplab}, all the pixels belonging to one instance share the same semantic label, which is usually easy to be identified by annotators. Differently, the human parsing task requires annotators to carefully distinguish those semantic parts of one person. Moreover, the situation will become even more challenging when the annotator got confused by the ambiguous boundaries between different semantic parts.

Due to these factors, there inevitably exists different types of label noises (as illustrated in Figure~\ref{fig:label-noise}) caused by the careless observations by annotators. This incomplete and low quality of the annotation labels will set a significant obstacle, which is usually ignored and prevents the performance of human parsing from increasing to a higher level. In this work, we investigate the problem of learning with noise in the human parsing task. Our target is to improve the model performance and generalization by progressively refining the noisy labels during the training stage.

In this paper, we introduce a purification strategy named Self-Correction for Human Parsing (SCHP), which can progressively promote the reliability of the supervised labels, as well as the learned models during the training process. Concretely, the whole SCHP pipeline can be divided into two sub-procedures, \textit{i.e.} the model aggregation and the label refinement procedure. Starting from a model trained on inaccurate annotations as initialization, we design a cyclically learning scheduler to infer more reliable pseudo-masks by iteratively aggregating the current learned model with the former optimal one in an online manner. Besides, those corrected labels can in turn to boost the model performance, simultaneously. In this way, the self-correction mechanism will enable the model or labels to mutually promote its counterpart, leading to a more robust model and accurate label masks as training goes on.

Besides, to tackle the problem of ambiguous boundaries between different semantic parts, we introduce a network architecture called Augmented Context Embedding with Edge Perceiving (A-CE2P). In principle, our network architecture is an intuitive generalization and augmentation of the CE2P~\cite{ruan2019devil} framework. We introduce a consistency constraint term to augment the CE2P, so that the edge information is not only implicitly facilitated the parsing result by feature map level fusion, but also explicitly constrained between the parsing and edge prediction. Note that we do not claim any novelty of our architecture structure, but only the superiority of the performance.

On the whole, our major contributions can be summarized as follows,
\begin{itemize}
    \item We propose a simple yet effective self-correction strategy SCHP for the human parsing task by online model aggregating and label refining which could mutually promote the model performance and label accuracy.
    \item We introduce a general architecture framework A-CE2P for human parsing that both implicitly and explicitly captures the boundary information along with the parsing information.
    \item We extensively investigate our SCHP on two popular human parsing benchmarks. Our method achieves the new state-of-the-art. In particular, we achieve the mIoU score of 59.36 on the large scale benchmark LIP, which outperforms the previous closest approach by 6.2 points.
\end{itemize}
\section{Related Work}\label{sec:related_work}
\textbf{Human Parsing.} Several different aspects of the human parsing task have been studied. Some early works~\cite{xia2017joint,liang2018look} utilized pose estimation together with the human parsing simultaneously as a multi-task learning problem. In \cite{ruan2019devil}, they cooperated the edge prediction with human parsing to accurately predict the boundary area. Most of the prior works assumed the fact that ground truth labels are correct and well-annotated. However, due to time and cost consuming, there inevitably exists lots of different label noises (as shown in Figure~\ref{fig:label-noise}). Meanwhile, it is impracticable to clean the pixel-level labels manually. Guided by this intuition, we try to tackle this problem via a novel self-correction mechanism in this paper.

\textbf{Pseudo-Labeling.} Pseudo-labeling~\cite{lee2013pseudo,reed2014training} is a typical technique used in semi-supervised learning. In semi-supervised learning setting, they assign pseudo-labels to the unlabeled data. However, in our fully supervised learning scheme, we are unable to locate the label noises, thus all ground truth labels are treated equally. From the perspective of distillation, the generated pseudo-label data contains much so-called \textit{dark knowledge}~\cite{hinton2015distilling} which could serve as a purification signal. Inspired by these findings, we design a cyclically learning scheduler to infer more reliable pseudo-masks by iteratively aggregating the current learned model with the former optimal one in an online manner. Also those corrected labels can in turn to boost the model performance, simultaneously.

\textbf{Self-Ensembling.} There is a line of researches~\cite{laine2016temporal,tarvainen2017mean,izmailov2018averaging} that exploit self-ensembling methods in various scenarios. For example, ~\cite{tarvainen2017mean} averaged model weights as self-ensembling and adopted in the semi-supervised learning task. In \cite{izmailov2018averaging}, they averaged the model weight and led to better generalization. Different from their method, our proposed self-correction approach is to correct the noisy training label via a model and label mutually promoting process. By an online manner, we average both model weights and the predictions simultaneously. To the best of our knowledge, we make a first attempt to formulate the label noise problem as the mutual model and label optimization in fine-grained semantic segmenting to boost the performance. Furthermore, our proposed method is online training with a cyclical learning scheduler and only exhaust little extra computation.

\section{Methodology}\label{sec:methodology}

\subsection{Revisiting CE2P}\label{sec:network}
CE2P~\cite{ruan2019devil} is a well-performing framework for the human parsing task. In the CE2P network, they cooperate the edge prediction with human parsing to accurately predict the boundary area. Concretely, CE2P consists of three key branches, \textit{i.e.} \textit{parsing} branch, \textit{edge} branch and \textit{fusion} branch. In particular, the \textit{edge} branch is employed to generate class-agnostic boundary maps. In the \textit{fusion} branch, both semantic-aware feature representations from the \textit{parsing} branch and the boundary-aware feature representations from the \textit{edge} branch are concatenated to further produce a refined human parsing prediction.

Although CE2P is a framework that has already incorporated the most useful functions from the semantic segmentation community. However, there are still some aspects that could be further strengthened. First, the conventional cross-entropy loss indirectly optimizes mean intersection-over-union (mIoU) metric, which is a crucial metric to reveal the comprehensive performance of the model. Second, CE2P only implicitly facilitates the parsing results with the edge predictions by feature-level fusion. There is no explicit constraint to ensure the parsing results maintaining the same geometry shape of the boundary predictions.

Moreover, few efforts have been made to investigate the versatility of the CE2P framework \textit{i.e.} the ability to accommodate other modules. Based on the key function, the \textit{parsing} branch can be divided into three modules, \textit{i.e.} backbone module, context encoding module and decoder module. Concretely, the backbone module could be plugged in with any fully-convolutional structure backbone such as ResNet-based~\cite{he2016deep} semantic segmentation network. The context encoding module utilizes the global context information to distinguish the fine-grained categories information. This module could be any effective context discovering module, \textit{e.g.} feature pyramid based approaches like PSP~\cite{zhao2017pyramid}, ASPP \cite{chen2017rethinking}, or attention-based modules like OCNet~\cite{yuan2018ocnet}. More detailed network architecture could refer to our code.

\begin{figure}[t]
\centering
\includegraphics[width=\linewidth]{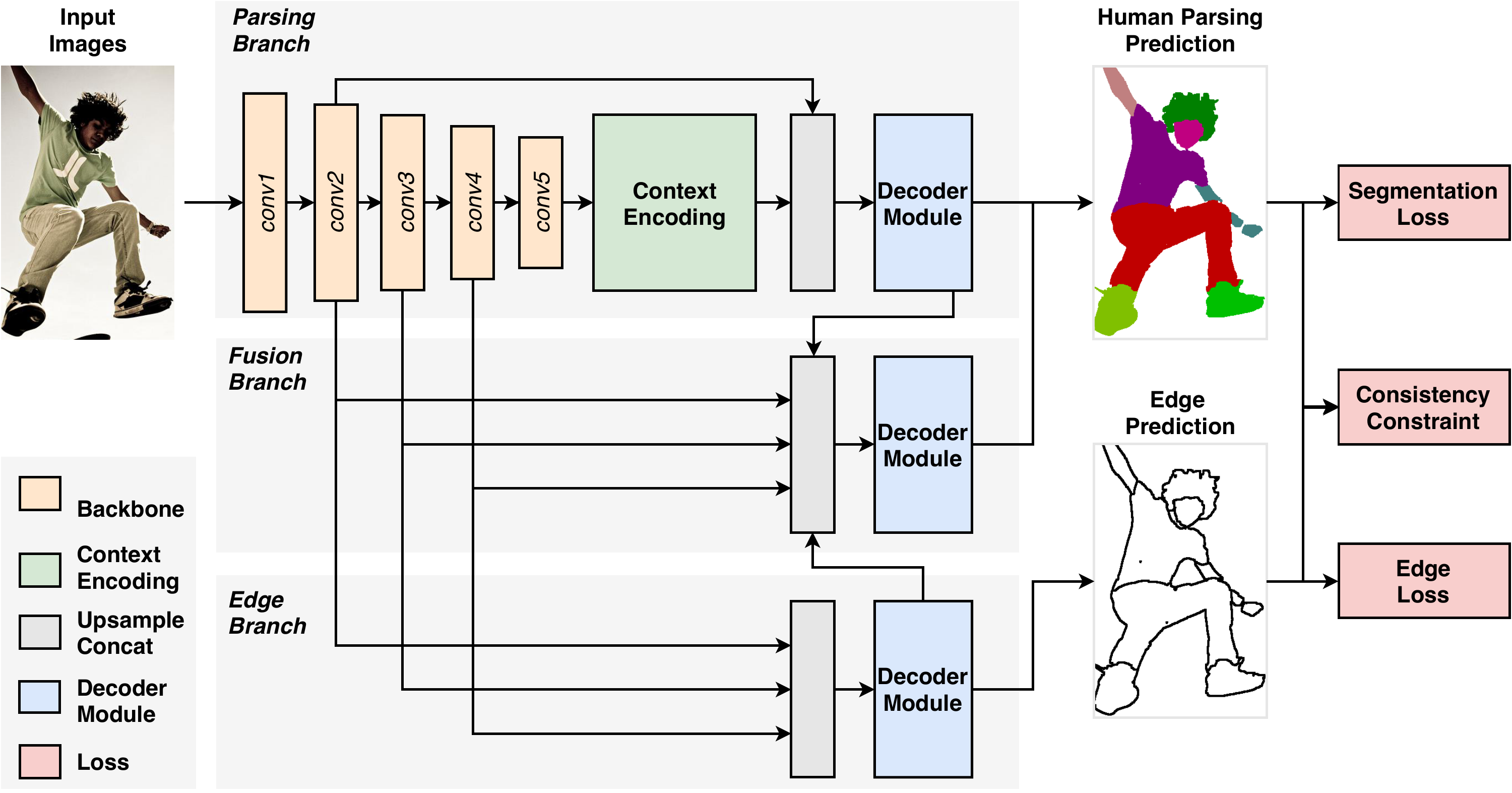} 
\caption{An overview of the Augmented-CE2P framework.}
\label{fig:schp-framework}   
\end{figure} 
\begin{figure*}[t]
\centering
\includegraphics[width=0.9\linewidth]{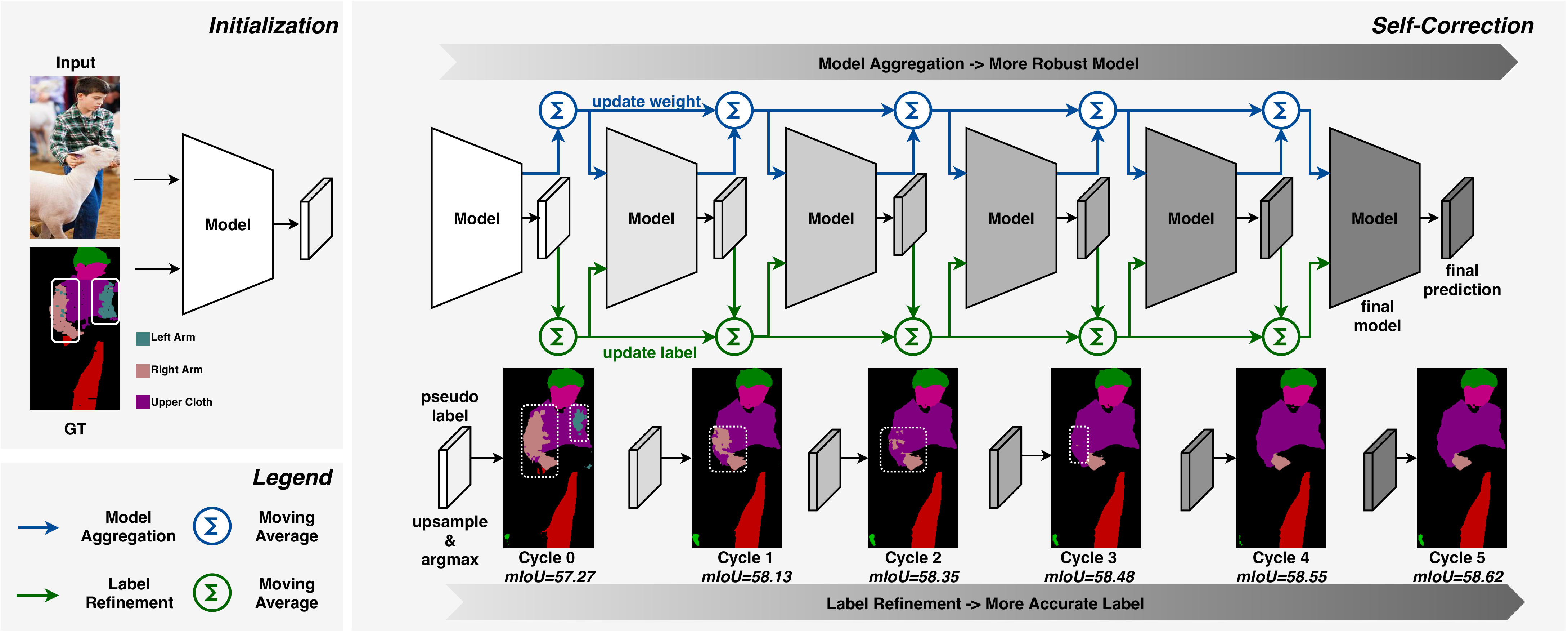} 
\caption{Illustration of the SCHP Pipeline. The self-correction mechanism will enable the model or labels to mutually promote its counterpart, leading to a more robust model and accurate label masks as training goes on. Label noises are specially marked with white boxes.}
\label{fig:self-correction}   
\end{figure*} 





\subsection{Augmented-CE2P}

The Augmented-CE2P (A-CE2P) is an intuitive generalization and augmentation of the CE2P, which can yield increased performance gain by augmenting additional powerful modules. In this work, our self-correction training employs the A-CE2P as the basic framework for conducting human parsing. We demonstrate the overview of the A-CE2P framework in Figure~\ref{fig:schp-framework}. Notably, several unique characteristics of A-CE2P are described as follows.

\textbf{Targeted Learning Objectives.}\label{sec:segmentation-objective-function}
For an image $I$, suppose the human parsing ground truth label is $\hat{y}_k^n$ and the parsing prediction is $y_k^n$, where n is the number of pixels for the k-th class. We define the pixel-level supervised objective using conventional cross-entropy loss as:
\begin{equation}
    \mathcal{L}_{cls}=-\frac{1}{N}\sum_k \sum_n {\hat{y}_k^n \log p(y_k^n)}.
\end{equation}
here $N$ is the number of pixels, $K$ is the number of classes.

It is known that conventional cross-entropy loss is usually convenient to train a neural network, but it facilitates mean intersection-over-union (mIoU) indirectly. To tackle this issue, following by \cite{berman2018lovasz}, we additionally introduce a tractable surrogate loss function for optimizing the mIoU directly. The final parsing loss function can be defined as a combination of the cross-entropy loss and the mIoU loss $\mathcal{L}_{miou}$,
\begin{equation}
    \mathcal{L}_{parsing} = \mathcal{L}_{cls} + \mathcal{L}_{miou}.
\end{equation}


\textbf{Consistency Constraint.}
In the CE2P, the balanced cross-entropy loss $\mathcal{L}_{edge}$ is adopted to optimize the edge prediction, so that the learned edge-aware features can help distinguish human parts and facilitate human parsing via the fusion branch indirectly. 

In the A-CE2P, we propose to further exploit the predicted boundary information by explicitly maintaining the consistency between the parsing prediction and the boundary prediction, \textit{i.e.} ensure that the predicted parsing result matches the predicted edge as exact as possible. Intuitively, we add a constraint term to penalized the mismatch:
\begin{equation}
    \mathcal{L}_{consistent} = \frac{1}{|N^+|}\sum_{n \in N^{+}}{|\Tilde{e}^n-e^n|},
\end{equation}
where $e^n$ is the edge maps predicted from the \textit{edge} branch and $\Tilde{e}^n$ is the edge maps generated from the parsing result $y_k^n$. To prevent the non-edge pixels dominate the loss, we only allow the positive edge pixels $n \in N^{+}$ for contributing the consistency constraint term.

In brief, the overall learning objective of our framework is
\begin{equation}
    \mathcal{L}=\lambda_1\mathcal{L}_{edge} + \lambda_2\mathcal{L}_{parsing} + \lambda_3\mathcal{L}_{consistent},
\label{eq:loss}
\end{equation}
where $\lambda_1, \lambda_2$ and $\lambda_3$ are hyper-parameters to control the contribution among these three losses. We jointly train the model in an end-to-end fashion by minimizing $\mathcal{L}$.

\subsection{Learning with Noise via Self-Correction}\label{sec:self-correction}
Based on the A-CE2P, we proposed the self-correction method that allows us to refine the label and get a robust model via an online mutual improvement process, illustrated in Figure~\ref{fig:self-correction}.

\textbf{Training Strategy.}\label{sec:training-strategy}
Our proposed self-correction training strategy is a model and label aggregating process, which can promote the model performance and refine the ground truth labels iteratively. This promotion relies on the initial performance of the model. In other words, if intermediate results generated by the network are not accurate enough, they may potentially harm the iteration process. Therefore, we start to run our proposed self-correction algorithm after a good initialization, \textit{i.e.} when the training loss starts to flatten with the original noisy labels. To make a fair comparison with other methods, we shorten the initial stage and keep the total training epochs as the same. After the initialization stage, we adopt a cyclically learning scheduler with warm restarts. Each cycle totally contains $T$ epochs. In practice, we use a cosine annealing learning rate scheduler with cyclical restart~\cite{loshchilov2016sgdr}. Formally, $\eta_{max}$ and $\eta_{min}$ are set to the initial learning rate and final learning rate, while $T_{cur}$ is the number of epochs since the last restart. Thus, the overall learning rate can be formulated as,
\begin{equation}
    \eta = \eta_{min} + \frac{1}{2}(\eta_{max} - \eta_{min})(1 +\cos(\frac{T_{cur}}{T}\pi)).
\label{eq:learning-rate}
\end{equation}

\textbf{Online Model Aggregation.} We aim to discover all the potential information from the past optimal models to improve the performance of the future model. In our cyclical training strategy, intuitively, the model will converge to a local-minimum at the end of each cycle. And there has great model disparity among these sub-optimal models. 
Here we denote the set of all the sub-optimal model we get after each cycle as $\Omega=\{\hat{\omega}_0,\hat{\omega}_1,...,\hat{\omega}_M\}$ and $M$ is the total number of cycles.

At the end of each cycle, we aggregate the current model weight $\hat{\omega}$ with the former sub-optimal one $\hat{\omega}_{m-1}$ to achieve a new model weight $\hat{\omega}_{m}$,
\begin{equation}
    \hat{\omega}_m = \frac{m}{m+1}\hat{\omega}_{m-1} + \frac{1}{m+1}\hat{\omega},
\label{eq:model-aggregation}
\end{equation}
where $m$ denotes the current cycle number and $0 \leq m \leq M$.

After updating the current model weight with the former optimal one from the last cycle, we forward all the training data for one epoch to re-estimate the statistics of the parameters (\textit{i.e.} moving average and standard deviation) in all batch normalization~\cite{ioffe2015batch} layers. During these successive cycles of model aggregation, the network leads to wider model optima as well as improved model's generalization ability.

\begin{algorithm}[t]
\caption{Self-Correction for Human Parsing}
\label{alg:malr}
\KwIn{Initialized model weight $\hat{\omega}_0$, original ground truth labels $\hat{y}_0$, cycle epoch length $T$, total number of iterations $M$}
\KwOut{Final network model $\hat{\omega}_M$}
Initialize the model weight $\hat{\omega} \gets \hat{\omega}_0$ \;
\For{$m \gets 1,2,...,M$}{  
\For{$t \gets 1,2,..T$}{
Update the learning rate $\eta$ by Eq.~\eqref{eq:learning-rate}\;
\For{each batch in training set}{
Calculate loss $\mathcal{L}$ by Eq.~\eqref{eq:loss} using $\hat{y}_{m-1}$\;
Gradient descending $\hat{\omega} \gets \hat{\omega}-\eta\nabla\mathcal{L}$\;
}
}
Model aggregation by Eq.~\eqref{eq:model-aggregation} to update $\hat{\omega}_m$\;
Re-calculate the BN layer parameters\;
Re-calculate the pseudo-mask $\hat{y}$ using $\hat{\omega}_m$\;
Label refinement by Eq.~\eqref{eq:label-aggregation} to update $\hat{y}_m$\;
}
\end{algorithm}
\begin{figure*}[t]
\centering
\includegraphics[width=\linewidth]{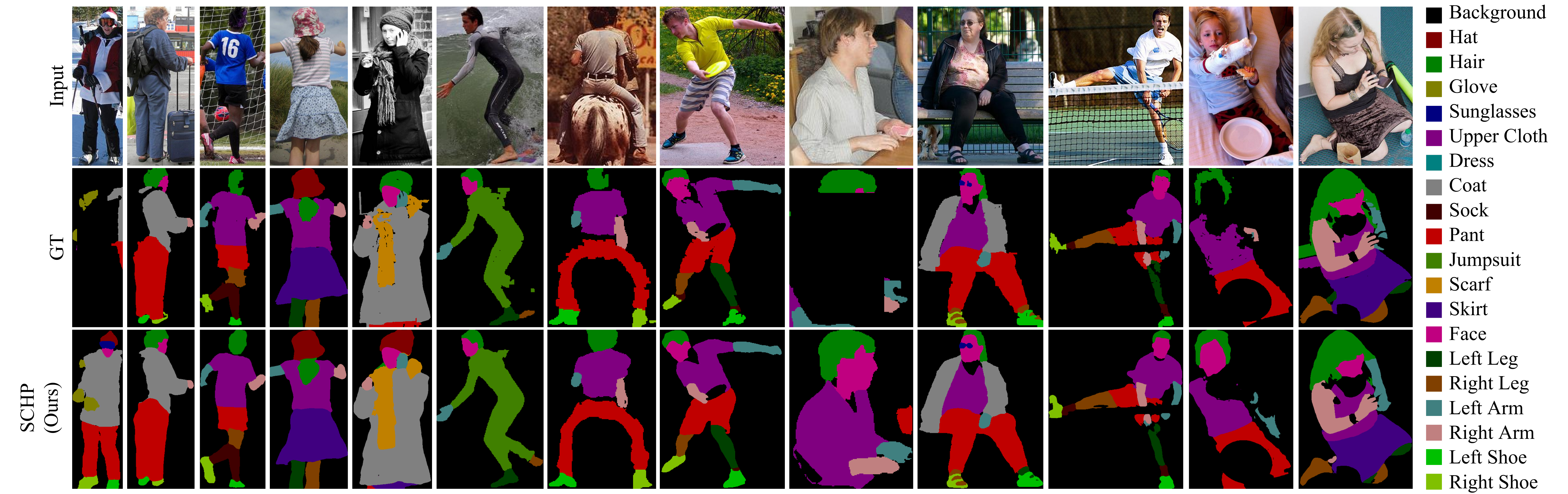} 
\caption{Visualization of SCHP results on LIP \texttt{validation} set. Note in most cases, our SCHP human parsing prediction is even better than the ground truth label. Zoom in to see details.}
\label{fig:lip-val-visualization}   
\end{figure*} 
\textbf{Online Label Refinement.}\label{sec:label-refinement} It is known that soft, multi-class labels may contain dark information~\cite{hinton2015distilling}. We aim to explore all these dark information to improve the performance and alleviate the label noises. 
After updating the model weight as mentioned in Eq.~\eqref{eq:model-aggregation}, we also update the ground truth of training labels. These generated pseudo-masks are more unambiguous, smooth and have the relational information between the fine-grain categories, which are taken as the supervised signal for the next cycle's optimization. During successive cycles of pseudo-label refinement, this improves the network performance as well as the generalization ability of the model. Also, these pseudo-masks potentially alleviate or eliminate the noise in the original ground truth. Here we denote the predicted label after each cycles as $Y=\{\hat{y}_0,\hat{y}_1,...,\hat{y}_M\}$. Same as the model weight averaging process, we update the ground truth label as follows,
\begin{equation}
    \hat{y}_m = \frac{m}{m+1}\hat{y}_{m-1} + \frac{1}{m+1}\hat{y},
\label{eq:label-aggregation}
\end{equation}
where $\hat{y}$ is the generated pseudo-mask by model $\hat{\omega}_m$. The detail of our proposed self-correction procedure is summarized in Algorithm~\ref{alg:malr}.

Note that the model and label are mutual improved step-by-step after each cyclical training process. The whole process is training in an online manner and does not need any extra training epochs. In addition, there is barely no extra computation required.

\subsection{Discussion}

\textbf{Can SCHP generalize to other tasks?} Our approach has no assumption for the data type. But the self-correction is base on the soft pseudo-label generated during the process. Thus our method could be applied in some other task such as classification and segmentation, but may be not applicable to regression tasks like detection. 

\textbf{Can SCHP still benefit with clean data?} Although we could achieve more performance gain with our proposed self-correction process on noisy datasets. However, when the ground truth is relatively clean, the online model aggregation process could serve as a self-model ensembling, which could lead to better performance and generalization. Still, the online label refinement process benefits from discovering the \textit{dark knowledge} using pseudo-mask instead of one-hot ground truth pixel-level label.

\textbf{Transduction \textit{vs.} Induction.}
In this work, we mainly focus on the supervised training scheme. Nevertheless, our approach can also be operated under semi-supervised learning manner, \textit{i.e.} we assume that all the test images are available at once and utilize all the test samples for self-correction process together with the training images jointly.

\section{Experiments}\label{sec:experiment_single}
In this section, we perform a comprehensive comparison of our SCHP with other single-person human parsing state-of-the-art methods along with thorough ablation experiments to demonstrate the contribution of each component.
\begin{table*}[t]
\centering
\footnotesize
\setlength{\tabcolsep}{1.25pt}
\resizebox{\linewidth}{!}{
\begin{tabular}{l|cccccccccccccccccccc|c}
\toprule
Method & hat & hair & glove & s-glass & u-clot & dress & coat & sock & pant & j-suit & scarf & skirt & face & l-arm & r-arm & l-leg & r-leg & l-shoe & r-shoe & bkg & mIoU  \\
\midrule
\midrule
Attention~\cite{chen2016attention} &58.87 &66.78 &23.32 &19.48 &63.20 &29.63 &49.70 &35.23 &66.04 &24.73 &12.84 &20.41 &70.58 &50.17 &54.03 &38.35 &37.70 &26.20 &27.09 &84.00 &42.92 \\
DeepLab~\cite{chen2017deeplab} &59.76 &66.22 &28.76 &23.91 &64.95 &33.68 &52.86 &37.67 &68.05 &26.15 &17.44 &25.23 &70.00 &50.42 &53.89 &39.36 &38.27 &26.95 &28.36 &84.09 &44.80 \\
SSL~\cite{gong2017look} &58.21 &67.17 &31.20&23.65&63.66 &28.31 &52.35&39.58 &69.40&28.61&13.70 &22.52&74.84 &52.83 &55.67 &48.22 &47.49 &31.80 &29.97 &84.64 &46.19 \\
MMAN \cite{luo2018macro} &57.66 &65.63 &30.07 &20.02 &64.15 &28.39 &51.98 &41.46 &71.03 &23.61 &9.65 &23.20 &69.54 &55.30 &58.13 &51.90 &52.17 &38.58 &39.05 &84.75 &46.81 \\
MuLA \cite{nie2018mutual} &- &- &- &- &- &- &- &- &- &- &- &- &- &- &- &- &- &- &- &- &49.30 \\
JPPNet \cite{liang2018look} & 63.55 & 70.20 & 36.16 & 23.48 & 68.15 & 31.42 & 55.65 & 44.56 & 72.19 & 28.39 & 18.76 & 25.14 & 73.36 & 61.97 & 63.88 & 58.21 & 57.99 & 44.02 & 44.09 & 86.26 & 51.37 \\
CE2P \cite{ruan2019devil} & 65.29 & 72.54 & 39.09 & 32.73 & 69.46 & 32.52 & 56.28 & 49.67 & 74.11 & 27.23 & 14.19 & 22.51 & 75.50 & 65.14 & 66.59 & 60.10 & 58.59 & 46.63 & 46.12 & 87.67 & 53.10 \\
\midrule
A-CE2P w/o SCHP & 69.59 & 73.02 & 45.21 & 35.59 & 69.85 &  35.97 & 56.96 & 51.06 & 75.79 & 30.41 & 22.00 & 27.07 & 75.79 & 68.54 & 70.30 & 67.83 & 66.90 & 53.53 & 54.08 & 88.11 & 56.88 \\
A-CE2P w/ SCHP & 69.96 & 73.55 & 50.46 & 40.72 & 69.93 & 39.02 & 57.45 & 54.27 & 76.01 & 32.88 & 26.29 & 31.68 & 76.19 & 68.65 & 70.92 & 67.28 & 66.56 & 55.76 & 56.50 & 88.36 & 58.62 \\
A-CE2P w/ SCHP\textsuperscript{\dag} & \bf{70.63} & \bf{74.09} & \bf{51.40}& \bf{41.70} & \bf{70.56} & \bf{40.06} & \bf{58.17} & \bf{55.17} & \bf{76.57} & \bf{33.78} & \bf{26.63} & \bf{32.83} & \bf{76.63} & \bf{69.33} & \bf{71.76} & \bf{67.93} & \bf{67.42} & \bf{56.56} & \bf{57.55} & \bf{88.40} & \bf{59.36} \\
\bottomrule
\end{tabular}
}
\caption{Comparison with state-of-the-arts on LIP \texttt{validation} set. \textsuperscript{\dag}{} designates the test time augmentation.}
\label{tab:lip-sota-comparsion}
\end{table*}
\begin{table}[t]
\centering
\scriptsize
\setlength{\tabcolsep}{0.8pt}
\resizebox{\linewidth}{!}{
\begin{tabular}{l|ccccccc|c}
\toprule
Method & head & torso & u-arm & l-arm & u-leg & l-leg & bkg & mIoU  \\
\midrule
\midrule
Attention~\cite{chen2016attention} &81.47 &59.06 &44.15 &42.50 &38.28 &35.62 &93.65 &56.39 \\
HAZN~\cite{xia2016zoom} &80.76 &60.50 &45.65 &43.11 &41.21 &37.74 &93.78 &57.54 \\
LG-LSTM~\cite{liang2016semantic} &82.72 &60.99 &45.40 &47.76 &42.33 &37.96 &88.63 &57.97 \\
SS-JPPNet~\cite{liang2018look} &83.26 &62.40 &47.80 &45.58 &42.32 &39.48 &94.68 &59.36 \\
MMAN~\cite{luo2018macro} &82.58 &62.83 &48.49 &47.37 &42.80 &40.40 &94.92 &59.91 \\
G-LSTM~\cite{liang2016graphlstm} &82.69 &62.68 &46.88 &47.71 &45.66 &40.93 &94.59 &60.16 \\
Part FCN~\cite{xia2017joint} &85.50 &67.87 &54.72 &54.30 &48.25 &44.76 &95.32 &64.39 \\
Deeplab~\cite{chen2017deeplab} &- &- &- &- &- &- &- & 64.94 \\
WSHP~\cite{fang2018weakly} &87.15 &72.28 &57.07 &56.21 &52.43 &50.36 &\textbf{97.72} &67.60 \\
PGN~\cite{Gong_2018_ECCV} &\textbf{90.89} &\textbf{75.12} &55.83 &64.61 &55.42 &41.57 &95.33 &68.40 \\
DPC~\cite{chen2018searching} &88.81 &74.54 &63.85 &63.73 &57.24 &54.55 & 96.66 &71.34 \\
\midrule
A-CE2P w/ SCHP &87.00 &72.27 &64.10 &63.44 &56.57 &55.00 &96.07 &70.63 \\
A-CE2P w/ SCHP\textsuperscript{\dag} &87.41 &73.80 &\textbf{64.98} &\textbf{64.70} &\textbf{57.43} &\textbf{55.62} &96.26 &\textbf{71.46} \\
\bottomrule
\end{tabular}
}
\caption{Comparison with state-of-the-arts on PASCAL-Person-Part \texttt{validation} set. \textsuperscript{\dag}{} designates the test time augmentation.}
\label{tab:pascal-sota-comparsion}
\end{table}

\subsection{Datasets and Evaluation Metrics}
\textbf{Datasets.}
We evaluate our proposed method on two single human parsing benchmarks, including LIP~\cite{liang2018look} and PASCAL-Person-Part~\cite{chen2014detect}. 

\emph{LIP}~\cite{liang2018look} is the largest human parsing dataset, which contains 50,462 images with elaborated pixel-wise annotations with 19 semantic human part labels. The images collected from the real-world scenarios contain human appearing with challenging poses and views, heavily occlusions, various appearances and low-resolutions. The datasets are divided images into 30,462 images for \texttt{train} set, 10,000 images for \texttt{validation} set and 10,000 for \texttt{test} set. 

\emph{PASCAL-Person-Part}~\cite{chen2014detect} is a relatively small dataset annotated from PASCAL VOC 2010, which contains 1,716 \texttt{train} images, 1,817 \texttt{validation} images. The ground truth label consists of six semantic parts including head, torso, upper/lower arms, upper/lower legs and one background class. This dataset is challenging due to large variations in scale. 

\textbf{Metrics.}
We report three standard metrics for the human parsing task, including pixel accuracy, mean accuracy, mean intersection over union (mIoU). Note the mIoU metric generally represents the overall parsing performance of the method. 
\begin{figure}[t]
\centering
\subfloat[Backbone.\label{fig:lip-backbone}]{\includegraphics[width=0.5\linewidth]{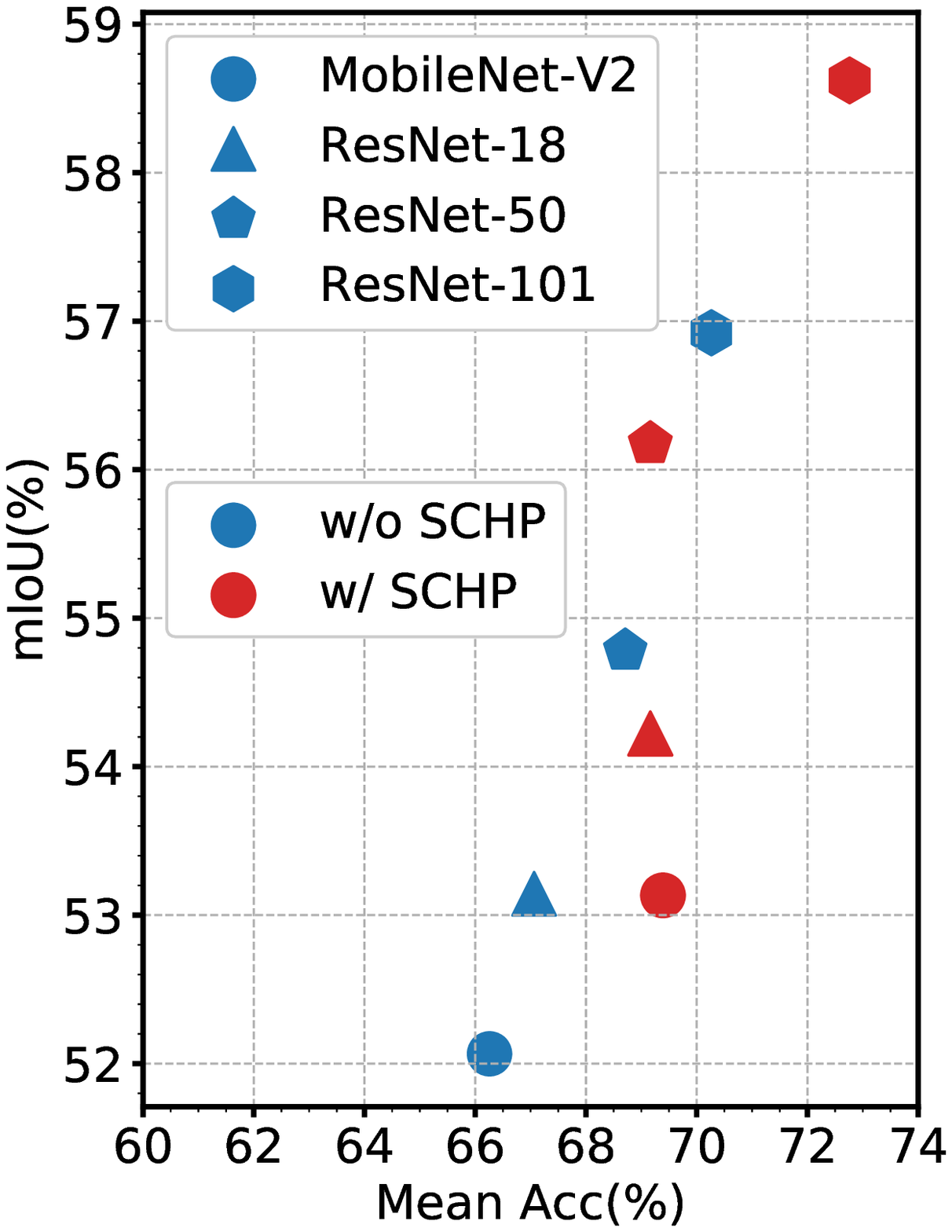}}
\subfloat[Context Encoding.\label{fig:lip-context-encoding}]{\includegraphics[width=0.5\linewidth]{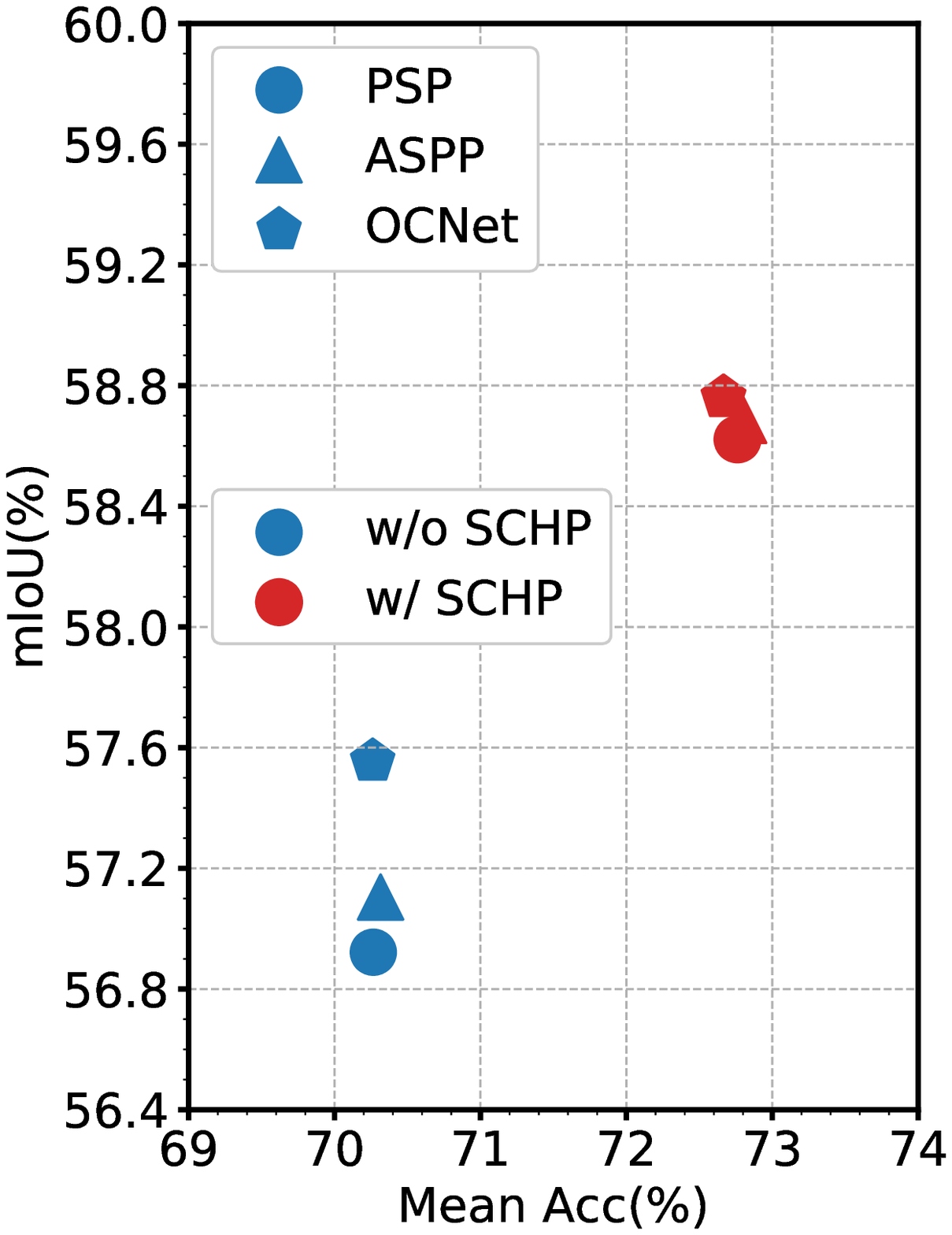}}
\caption{Effect of SCHP with different backbones and context encoding modules. All experiments are conducted on LIP \texttt{validation} set.}
\label{fig:lip-generic-framework}   
\end{figure} 

\textbf{Implementation Details.} We choose the ResNet-101~\cite{he2016deep} as the backbone of the feature extractor and use an ImageNet~\cite{deng2009imagenet} pre-trained weights. Specifically, we fix the first three residual layers and set the stride size of last residual layer to 1 with a dilation rate of 2. In this way, the final output is enlarged to 1/16 resolution size \textit{w.r.t} the original image. We adopt pyramid scene parsing network~\cite{zhao2017pyramid} as the context encoding module. We use $473\times473$ as the input resolution. Training is done with a total batch size of 36. For our joint loss function, we set the weight of each term as $\lambda_1=1,\lambda_2=1,\lambda_3=0.1$. The initial learning rate is set as 7e-3 with a linear increasing warm-up strategy for 10 epochs. We train our network for 150 epochs in total for a fair comparison, the first 100 epochs as initialization following 5 cycles each contains 10 epochs of the self-correction process.
\begin{table}[t]
\centering
\footnotesize
\begin{tabular}{ccc|ccc}
\toprule
\multicolumn{3}{c|}{Loss} & \multirow{2}{*}{Pixel Acc} & \multirow{2}{*}{Mean Acc} & \multirow{2}{*}{mIoU} \\
E & I & C &  &  &  \\
\midrule\midrule
- & - & - & 86.93 & 65.12 & 53.75 \\
\checkmark & - & - & 87.51 & 65.42 & 54.14 \\
\checkmark & \checkmark & - & 87.57 & 68.20 & 56.33 \\
\checkmark & \checkmark & \checkmark & 87.68 & 68.79 & 56.88 \\
\bottomrule
\end{tabular}
\caption{Each component of our loss function is evaluated on LIP \texttt{validation} set, including edge loss (E), IoU loss (I) and consistency constraint (C).}
\label{tab:ablation-lip-loss}
\end{table}
\begin{table}[t]
\centering
\footnotesize
\begin{tabular}{cc|ccc}
\toprule
\multicolumn{2}{c|}{Method} & \multirow{2}{*}{Pixel Acc} & \multirow{2}{*}{Mean Acc} & \multirow{2}{*}{mIoU} \\
MA & LR &  &  &  \\
\midrule\midrule
- & - & 87.68 & 68.79 & 56.88 \\
\checkmark & - & 87.90 & 71.27 & 57.94 \\
- & \checkmark & 87.86 & 70.88 & 57.44 \\
\checkmark & \checkmark & 88.10 & 72.76 & 58.62 \\
\bottomrule
\end{tabular}
\caption{The effect of our proposed model aggregation (MA) and label refinement (LR) strategy is evaluated on LIP \texttt{validation} set.}
\label{tab:ablation-lip-mala}
\end{table}
\subsection{Comparison with state-of-the-arts}
In Table~\ref{tab:lip-sota-comparsion}, we compare the performance of our network with other state-of-the-art methods on the LIP. It can be observed that even our baseline model outperforms all the other state-of-the-art methods, which illustrates the effectiveness of the A-CE2P framework. In particular, we also apply test-time augmentation with multi-scale and horizontal flipping to make a fair comparison with others. Our SCHP outperforms the others with a large gain, achieving mIoU improvement by 6.26\%, which is a significant boost considering the performance at this level. Our proposed approach achieves a large gain especially for some categories with few samples like \textit{scarf, sunglasses} and some confusing categories such as \textit{dress, skirt} and left-right confusion. The gains are mainly from using both model aggregation and label refinement for our self-correction process. Furthermore, the qualitative comparison between the predicted results of SCHP and ground truth annotations is shown in Figure~\ref{fig:lip-val-visualization}. We can see that our SCHP can achieve even better parsing results than the ground truth ones.


To validate the generalization ability of our method, we further report the comparison with other state-of-the-arts on PASCAL-Person-Part in Table~\ref{tab:pascal-sota-comparsion}. It can be observed that our SCHP outperforms all the previous approaches. Particularly, our SCHP beats the DPC~\cite{chen2018searching}. DPC is a network architecture search (NAS)-based method which easily achieves optimal results on the small dataset. Besides, instead of using more powerful backbone models such as Xception~\cite{chollet2017xception} in DPC, we only adopt ResNet-101 as the backbone. In addition, DPC leverages MS COCO~\cite{lin2014microsoft} as additional data for pre-training, while our model is only pre-trained on ImageNet. All these results well demonstrate the superiority and generalization of our proposed approach.


\subsection{Ablation Study}
We perform extensive ablation experiments to illustrate the effect of each component in our SCHP. All experiments are conducted on LIP benchmark.

\begin{figure}[t]
\centering
\includegraphics[width=\linewidth]{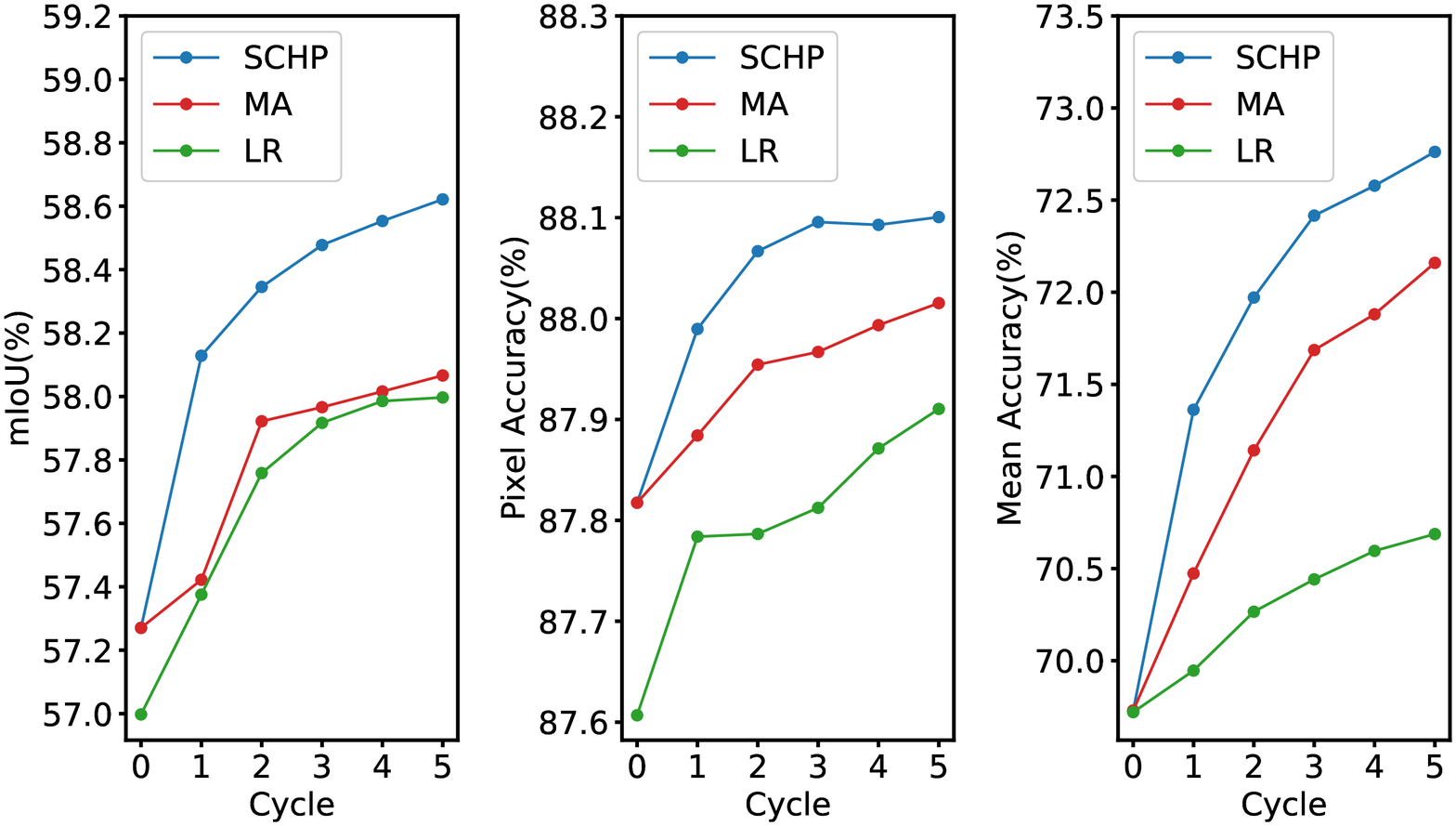}
\caption{Performance curves \textit{w.r.t} different training cycles. All experiments are conducted on LIP \texttt{validation} set. The mIoU, pixel accuracy, mean accuracy are reported, respectively.}
\label{fig:lip-malr-cycle}   
\end{figure} 
\textbf{Alternatives Architectures.}
Since our proposed SHCP is a generic framework, we could merely plug-and-play with various backbones and context encoding modules. Figure~\ref{fig:lip-backbone} shows SHCP with different backbones from lightweight model MobileNet-V2~\cite{sandler2018mobilenetv2} to relatively heavy backbone ResNet-101~\cite{he2016deep}. Interestingly, the lightweight MobileNet-V2 achieves the mIoU score of 52.06, with the benefit of SCHP leads to 53.13. This result is even better than some previous results~\cite{ruan2019devil} achieved by ResNet-101. We note that deeper network (18 \textit{vs.} 50 \textit{vs.} 101) tends to perform better. Regardless of different backbones, our SCHP consistently brings consistent positive gains, 1.08, 1.40 and 1.70, respectively in terms of mIoU. It suggests the robustness of our proposed method. As shown in Figure~\ref{fig:lip-context-encoding}, we further examine the robustness of our SCHP by varying the context encoding module. In particular, we choose three different types of modules, including multi-level global average pooling based module pyramid scene parsing network (PSP)~\cite{zhao2017pyramid}, multi-level atrous convolutional pooling based module atrous spatial pyramid pooling (ASPP)~\cite{chen2017rethinking} and attention-mechanism based module OCNet~\cite{yuan2018ocnet}. Despite the similar basic performance of these three modules, our SCHP unfailingly obtains mIoU increased by 1.70, 1.30, 1.00 points for PSP, ASPP and OCNet respectively. This further highlights the effectiveness of self-correction mechanism in our approach. Note that although we could achieve even better results with these advanced modules, the network structure modification is not the focus of this study. In our baseline model, we choose to use ResNet-101 as backbone and PSP as context encoding module.

\textbf{Influence of Learning Objectives.}
Our network is trained in an end-to-end manner with composite learning objectives describes as Eq.~\eqref{eq:loss}. An evaluation of different learning objectives is shown in Table~\ref{tab:ablation-lip-loss}. In this table, \textit{E} denotes the binary cross-entropy loss to optimize the boundary prediction. \textit{I} denotes the tractable surrogate function optimizing the mIoU metric. \textit{C} denotes the consistency constraint term for maintain the consistency between parsing result and boundary result. Without all these three terms, only the basic cross-entropy loss function for parsing takes effect. By introducing the edge information, the performance improves mIoU by about 0.4. This gain is mainly due to the accurate prediction at the boundary area between semantic parts. Additionally, we compare the result further adding the IoU loss. As can be seen, the IoU loss significantly boosts the mean accuracy by 2.8 points and mIoU by 2.2 points, but the pixel accuracy almost remains the same level. This highlights that IoU loss largely resolves the challenge of prediction accuracy especially at some small area and infrequent categories. Finally, the result shows a gain of 0.59 mean accuracy and 0.55 mIoU when applying the consistency between parsing segmentation and edge prediction.
\begin{figure}[t]
\centering
\includegraphics[width=\linewidth]{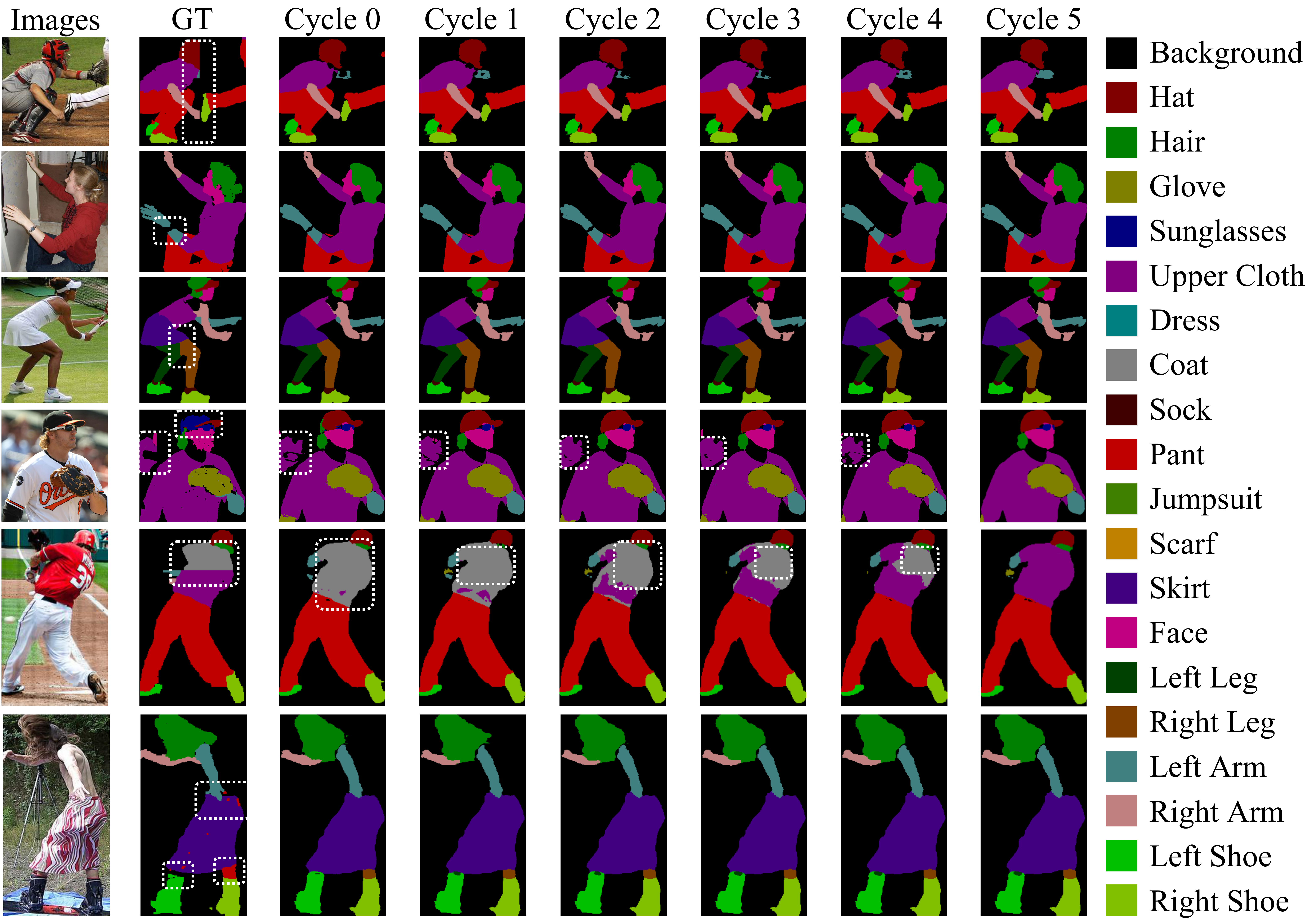}
\caption{Visualization of our self-correction process in LIP \texttt{train} set. Label noises are emphasized with white dotted boxes. Better zoom in to see the details.}
\label{fig:lip-self-correction-extra}   
\end{figure} 
\textbf{Effect of Self-Correction.}
In Table~\ref{tab:ablation-lip-mala}, we validate the effect of each component in our proposed SCHP, including the model aggregation (MA) process and the label refinement (LR) process. All experiments are conducted upon with A-CE2P framework. When there are no MA and LR involved, our method reduces to the conventional training process. By employing the MA process, the result shows a gain of mIoU by 1.1 points. While only benefit from the LR process, we achieve 0.6 point improvement. We achieve the best performance by simultaneously introducing these two processes. We can observe that the model aggregation and label refinement mutually promote each other in our SCHP.

To better qualitatively describe our SCHP, Figure~\ref{fig:lip-self-correction} shows the visualization of the generated pseudo-masks during the self-correction cycles. Note that all these pseudo-masks are up-sampled to the original size and applied argmax operation for better illustration. Label noises like inaccurate boundary, confused fine-grained categories, confused mirror categories, multi-person occlusion are alleviated and partly resolved during the self-correction cycles. Unsurprisingly, some of the boundaries of our corrected labels are prone to be more smooth than the ground truth labels. All these results demonstrate the effectiveness of our proposed self-correction method. Intuitively, our self-correction process is a mutual promoting process benefiting both model aggregation and label refinement. During the self-correction cycles, the model gets increasingly more robust, while by exploring the dark information from pseudo-masks produced by the enhanced model, the label noises are corrected in an implicit manner. The fact that corrected labels are smooth than the ground truth also illustrates the effectiveness of our model architecture design for combining the edge information.

\textbf{Influence of Self-Correction Cycles.}
We achieve the goal of self-correction by a cyclically learning scheduler. The number of cycles is a virtual hyper-parameter for this process. To make a fair comparison with other methods~\cite{ruan2019devil}, we maintain the entire training epoch unchanged. The performance curves are shown in Figure~\ref{fig:lip-malr-cycle}. It is evident that the performance consistently improves during the process, with the largest improvement after the first cycle and tendency saturates at the end. Our method may achieve even higher performance when extending more training epochs. It is noteworthy the performance of \textit{MA}, \textit{LR} and \textit{SCHP} is not same at cycle 0. This small gap is caused due to the re-estimation of BatchNorm parameters. From the performance curve, We also intelligibly demonstrate the mutual benefit of the model aggregation and the label refinement process. More visualization of the self-correction process is illustrated in Figure \ref{fig:lip-self-correction-extra}.

\section{Conclusion and Future Work} \label{sec:conclusion}
In this paper, we propose an effective self-correction strategy to deal with the label noises for the human parsing task. Our proposed method achieves the new state-of-the-art with a large margin gain. Moreover, the self-correction mechanism is a general strategy for training and can be incorporated into any frameworks to make further performance improvement. In the future, we would like to extend our method to multiple-person human parsing and video multiple-person human parsing tasks.

\section{Acknowledgement}
We thank Ting Liu and Tao Ruan for providing insights and expertise to improve this work.

{\small
\bibliographystyle{ieee_fullname}
\bibliography{schp}
}

\end{document}